\documentclass[runningheads]{llncs}
\usepackage{graphicx}
\usepackage{url}
\usepackage{amsmath,amssymb} % define this before the line numbering.
\usepackage{color}
\usepackage[width=122mm,left=12mm,paperwidth=146mm,height=193mm,top=12mm,paperheight=217mm]{geometry}
\begin{document}
% \renewcommand\thelinenumber{\color[rgb]{0.2,0.5,0.8}\normalfont\sffamily\scriptsize\arabic{linenumber}\color[rgb]{0,0,0}}
% \renewcommand\makeLineNumber {\hss\thelinenumber\ \hspace{6mm} \rlap{\hskip\textwidth\ \hspace{6.5mm}\thelinenumber}}
% \linenumbers
\pagestyle{headings}
\mainmatter
\def\ECCV18SubNumber{***}  % Insert your submission number here
\newcommand{\footnoteref}[1]{\textsuperscript{\ref{#1}}}

\title{Efficient Facial Representations for Age, Gender and Identity Recognition in Organizing Photo Albums using Multi-output CNN} % Replace with your title

\titlerunning{Efficient Facial Representations for Age, Gender and Identity Recognition}

\authorrunning{A.V. Savchenko}

\author{Andrey V. Savchenko}
\institute{\textsuperscript{1}Samsung-PDMI Joint AI Center, St. Petersburg Department of Steklov Institute of Mathematics \\ \textsuperscript{2}National Research University Higher School of Economics\\
Nizhny Novgorod, Russia}

\maketitle

\begin{abstract}
This paper is focused on the automatic extraction of persons and their attributes (gender, year of born) from album of photos and videos. We propose the two-stage approach, in which, firstly, the convolutional neural network simultaneously predicts age/gender from all photos and additionally extracts facial representations suitable for face identification. We modified the MobileNet, which is preliminarily trained to perform face recognition, in order to additionally recognize age and gender. In the second stage of our approach, extracted faces are grouped using hierarchical agglomerative clustering techniques. The born year and gender of a person in each cluster are estimated using aggregation of predictions for individual photos. We experimentally demonstrated that our facial clustering quality is competitive with the state-of-the-art neural networks, though our implementation is much computationally cheaper. Moreover, our approach is characterized by more accurate video-based age/gender recognition when compared to the publicly available models. 
\end{abstract}

\section{Introduction}
Nowadays, due to the extreme increase in multimedia resources there is an urgent need to develop intelligent methods to process and organize them~\cite{manju2015organizing}. For example, the task of automatic organizing photo and video albums is attracting increasing attention~\cite{sokolova2017organizing,zhang2002hierarchical}. The various photo organizing systems allow users to group and tag photos and videos in order to retrieve large number of images in the media library~\cite{GFW}. The most typical processing of a gallery includes the face grouping, and each group can be automatically tagged with the facial attributes, i.e., age (born year) and gender~\cite{eidinger2014age}. Hence, the task of this paper is formulated as follows: given a large number of unlabeled facial images, cluster the images into individual persons (identities)~\cite{GFW} and predict age and gender of each person~\cite{rothe2015dex}. 

This problem is usually solved using deep convolutional neural networks (CNNs)~\cite{goodfellow2016deep}. At first, the clustering of photos and videos that contains the same person is performed using the known face verification~\cite{crosswhite2017template,wang2018additive} and identification~\cite{savchenko2018unconstrained} methods. The age and gender of extracted faces can be recognized by other CNNs~\cite{eidinger2014age,rothe2015dex}. Though such approach works rather well, it requires at least three different CNNs, which increases the processing time, especially if the gallery should be organized on mobile platforms in offline mode. Moreover, every CNN learns its own face representation, which quality can be limited by the small size of the training set or the noise in the training data. The latter issue is especially crucial for age prediction, which contains incorrect ground truth values of age.

It is rather obvious that the closeness among the facial processing tasks can be exploited in order to learn efficient face representations which boosts up their individual performances. For instance, simultaneous face detection, landmark localization, pose estimation, and gender recognition is implemented in the paper~\cite{ranjan2017hyperface} by a single CNN. 

Therefore the goal of our research is to improve the efficiency in facial clustering and age and gender prediction by learning face representation using preliminarily training on domain of unconstrained face identification from very large database. In this paper we specially developed a multi-output extension of the MobileNet~\cite{howard2017mobilenets}, which is pre-trained to perform face recognition using the VGGFace2 dataset~\cite{cao2018vggface2}. Additional layers of our network are fine-tuned for age and gender recognition on Adience~\cite{eidinger2014age} and IMDB-Wiki~\cite{rothe2015dex} datasets. Finally, we propose a novel approach to face grouping, which deals with several challenges of processing of real-world photo and video albums.

The rest of the paper is organized as follows: in Section~\ref{sect:materials}, we formulate the proposed approach of organizing facial photos with simultaneous prediction of age and gender of obtained persons. In Section~\ref{sect:exper}, we present the experimental results in face clustering for the LFW, Gallagher and GFW datasets and the video-based age/gender recognition for video clips from Eurecom Kinect, Indian Movie, EmotiW and IJB-A. Finally, concluding comments are given in Section~\ref{sect:conclusion}.

\section{Materials and Methods}
\label{sect:materials}

\subsection{Multi-output CNN for Simulatenous Age, Gender and Identity Recognition}
\label{sect:convnet}

In this paper we consider several different facial analytic tasks. We assume that the facial regions are obtained in each image using any appropriate face detector, e.g., either traditional multi-view cascade Viola-Jones classifier or more accurate CNN-based methods~\cite{zhang2016joint}. The \textit{gender} recognition task is a binary \textit{classification} problem, in which the obtained facial image is assigned to one of two classes (male and female). The \textit{age} prediction is the special case of \textit{regression} problem, though sometimes it is considered as a multi-class classification with, e.g., $N=100$ different classes, so that it is required to predict in an observed person is 1,2,... or 100 years old~\cite{rothe2015dex}. In such case these two tasks become very similar and can be solved by traditional deep learning techniques. Namely, the large facial dataset of persons with known age and/or gender is gathered, e.g. the IMDB-Wiki from the paper~\cite{rothe2015dex}. After that the deep CNN is learned to solve the classification task. The resulted networks can be applied to predict age and gender given a new facial image.

The last task examined in this paper, namely, unconstrained \textit{face identification} significantly differs from age and gender recognition. We consider the unsupervised learning case, in which facial images акщь a gallery set should be assigned to one of $C\ge1$ subjects (identities). The number of subjects $C$ is generally unknown. The training sample is usually rather small (we can assume that $C \approx R$) to train complex classifier (e.g. deep CNN). Hence, the domain adaptation can be applied~\cite{goodfellow2016deep}: each image is described with the off-the-shelf feature vector using the deep CNN, which has been preliminarily trained for the \textit{supervised face identification} from large dataset, e.g., CASIA-WebFace, VGGFace/VGGFace2 or MS-Celeb-1M. The $L_2$-normalized outputs at the one of last layers of this CNN for each $r$-th gallery image are used as the $D$-dimensional feature vectors $\mathbf{x}_r=[x_{r;1},...,x_{r;D}]$. Finally, any appropriate clustering method, i.e., hierarchical agglomerative clustering~\cite{aggarwal2013data}, can be used to make a final decision for these feature vectors.

\begin{figure}
\centering
\includegraphics[width=0.6\linewidth]{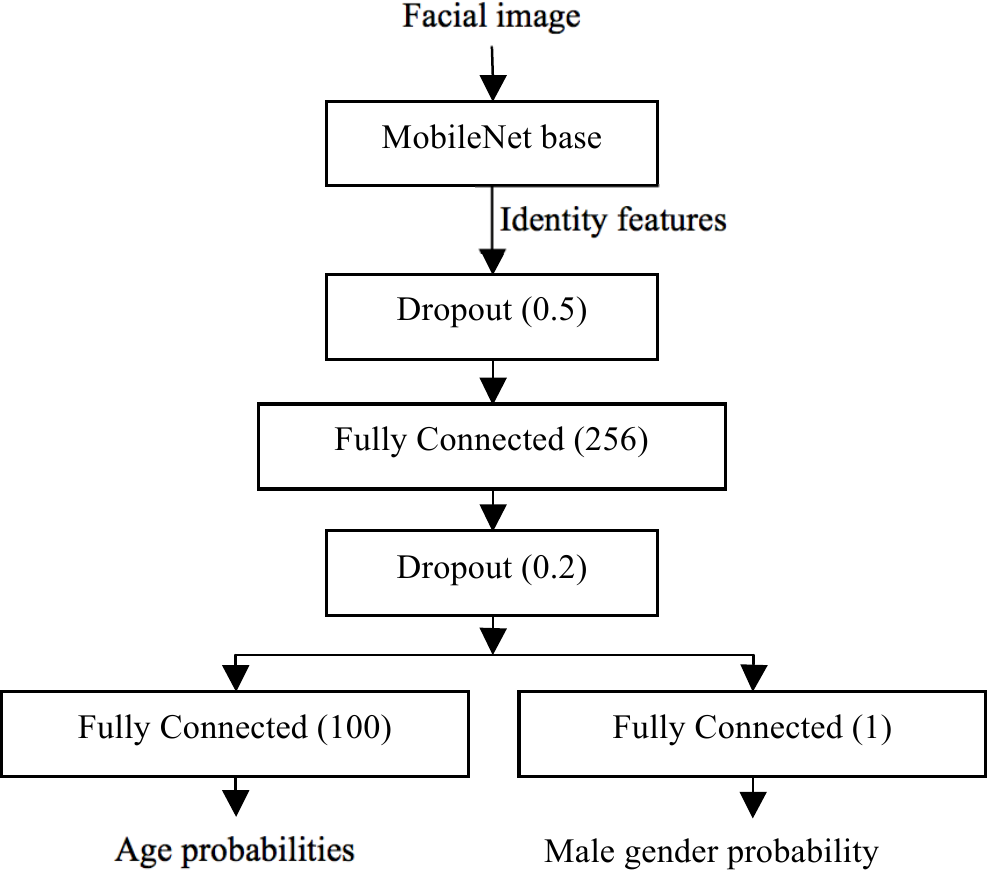}
\caption{Our CNN}
\label{fig:1}
\end{figure}

In most research studies all these tasks are solved by independent CNNs even though it is necessary to solve all of them. As a result, the processing of each facial image becomes time-consuming, especially for offline mobile applications. In this paper we propose to solve all these tasks by the same CNN. In particular, we assume that the features extracted during face identification can be rather rich for any facial analysis. For example, it was shown the the VGGFace features~\cite{parkhi2015deep} can be used to increase the accuracy of visual emotion recognition~\cite{kaya2017video,rassadin2017group}. As our main requirement is the possibility to use the CNN on mobile platforms, we decided to use straightforward modification of the MobileNet~\cite{howard2017mobilenets} (Fig.~\ref{fig:1}). The bottom of our network (conventional MobileNet v1 pre-trained on ImageNet) extracts the representations suitable for face identification. It was experimentally noticed that one new hidden layer with dropout regularization after extraction of identity features slightly improves the power of age and gender recognition performed by two independent fully connected layers with softmax and sigmoid outputs, respectively.

The learning of our model is performed incrementally, At first, we train the base MobileNet for face identification using very large dataset, e.g., VGGFace2 with 3M photos of 10K subjects~\cite{cao2018vggface2}. Next, the last classification layer is removed, and the weights of the MobileNet base are freezed. Finally, the remaining layers in the head are learned for age and gender recognition. In our study we populate the training dataset by 300K frontal cropped facial images from the IMDB-Wiki dataset~\cite{rothe2015dex}. Unfortunately, the age groups in this dataset are very imbalanced, so the trained models work incorrectly for faces of very young or old people. Hence, we decided to add all (15K) images from the Adience~\cite{eidinger2014age} dataset. As the latter contains only age intervals, e.g., ``(0-2)", ``(60-100)", we put all images from this interval to the middle age, e.g., ``1" or ``80".

It is necessary to emphasize that not all images in the IMDB-Wiki contains information about both age and gender. Moreover, the gender is sometimes unknown in the Adience data. As a result, the number of faces with both age and gender information is several times smaller when compared to the whole number of facial images. Finally, the gender data for different ages is also very imbalanced. Thus, we decided to train both heads of the CNN (Fig.~\ref{fig:1}) independently using different training data for age and gender classification. In particular,  we alternate the mini-batches with age and gender info, and train only the part of our network, i.e., the weights of the fully connected layer in the age head of our model are not updated for the mini-batch with the gender info.

This CNN has the following advantages. First of all, it is obviously very efficient due to either usage of the MobileNet base or the possibility to simultaneously solve all three tasks (age, gender and identity recognition) without need to implement an inference in three different networks. Secondly, in contrast to the publicly available datasets for age and gender prediction, which are rather small and dirty, our model exploit the  potential of very large and clean face identification datasets to learn very good face representation. Moreover, the hidden layer between the identity features and two outputs further combines the knowledge necessary to predict age and gender. As a result, our model makes it possible to increase the accuracy of age/gender recognition when compared to the models trained only on specific datasets, e.g. IMDB-Wiki or Adience. Subsection~\ref{sect:AgeGen} will experimentally support this claim.

\subsection{Proposed Pipeline for Organizing Photo and Video Albums}
\label{sect:pipeline}

\begin{figure}
\centering
\includegraphics[width=\linewidth]{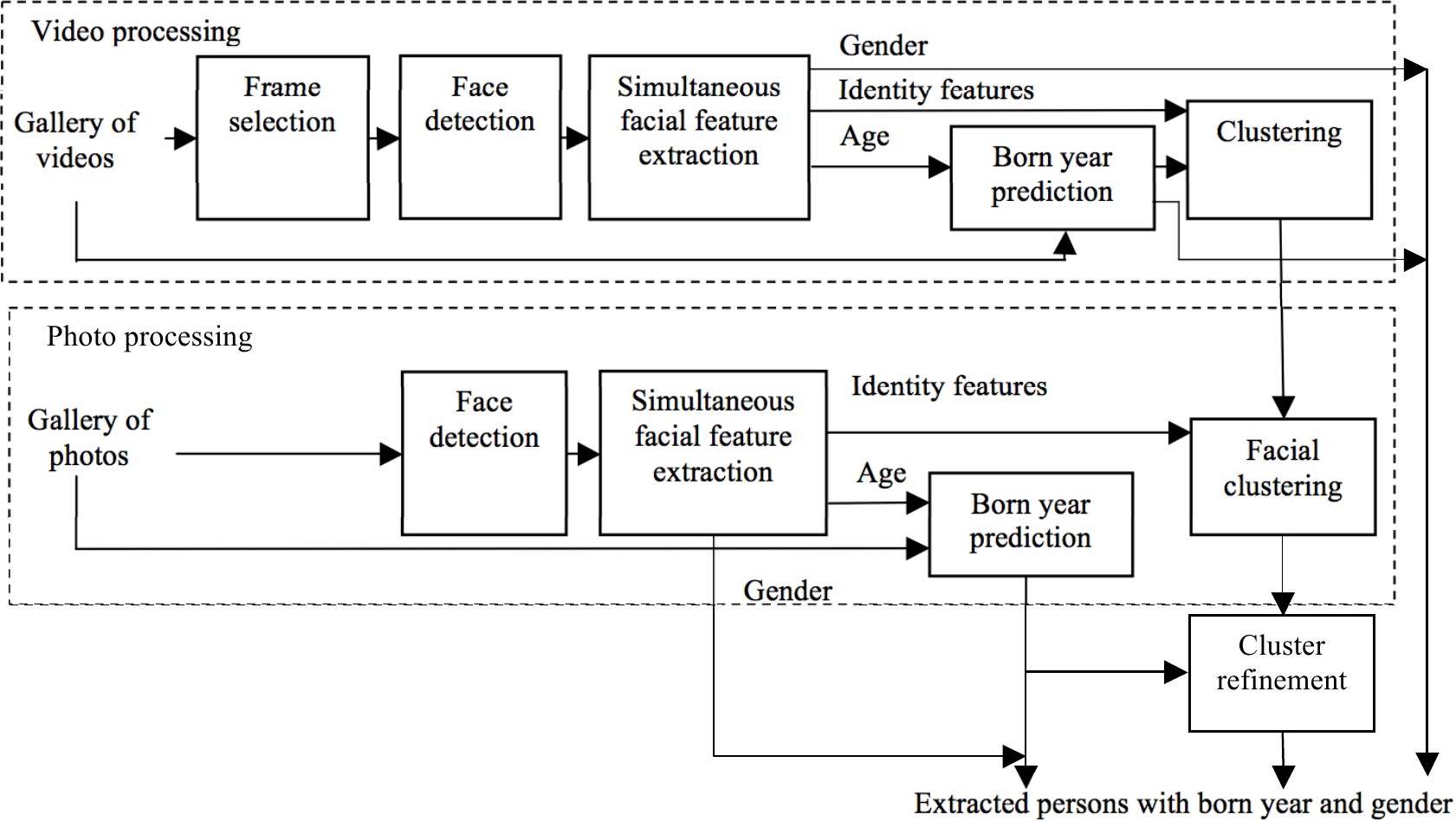}
\caption{Proposed pipeline}
\label{fig:2}
\end{figure}

The complete data flow of the usage of the CNN (Fig.~\ref{fig:1}) for organizing albums with photos and videos  the is presented in Fig.~\ref{fig:2}.

Here we detect faces in each photo using the MTCNN. Next, an inference in our CNN is performed for all detected faces $X_r$ in order to extract $D$ identity features and predict age and gender. After that, all facial identity feature vectors are clustered. As the number of subjects in the photo albums is usually unknown, we used a hierarchical agglomerative clustering~\cite{aggarwal2013data}. Only rather large clusters with a minimal number of faces are retained during the cluster refinement. The gender and the born year of a person in each cluster are estimated by appropriate fusion techniques, e.g., simple voting or maximizing the average posterior probabilities at the output of the CNN (Fig.~\ref{fig:1}). For example, the product rule~\cite{kittler1998combining} can be applied is we naively assume the independence of all facial images $X_r, r \in \{r_1,...,r_M\}$ in a cluster:

\begin{align}
  \underset{n \in \{1,...,N\}}{\max} \prod_{m=1}^{M} {p_n(X_{r_m})} =\underset{n \in \{1,...,N\}}{\max} \sum_{m=1}^{M} {\log p_n(X_{r_m})},
\end{align}
where $N$ is the total number of classes (in our study $N=2$ and $N=100$ for gender and age recognition, respectively) and $p_n(X_{r_m})$ is the $n$-th output of the CNN for the input image $X_{r_m}$.

The same procedure is repeated for all video files. We select only each of, e.g., 3 or 5 frames, in each video clip, extract identity features of all detected faces and initially cluster \textit{only} the faces found in this clip. After that we compute the normalized average of identity features of all clusters~\cite{sokolova2017organizing}, and add them to the dataset $\{X_r\}$ so that the ``Facial clustering" module handles both features of all photos and average feature vectors of subjects found in all videos.

Let us summarize the main novel parts of this pipeline despite the usage of our multi-output neural network (Fig.~\ref{fig:1}).

Firstly, the simple age prediction by maximizing the output of the corresponding head of our CNN is not accurate due to the imbalance of our training set. Addition of the Adience data leads to the decision in favour of one of the majority class. Hence, we decided to aggregate the outputs $\{p_a(X_r)\}$ of the age head. However, we experimentally found that the fusion of \textit{all} outputs is again inaccurate, because the majority of subjects in our training set are 20-40 years old. Thus, we propose to choose only $L\in\{1,2,...,100\}$ indices $\{a_1, ...,a_L\}$ of the maximal outputs and compute the expected mean $\overline{a}(X_r)$ for each gallery facial image $X_r$ using normalized top outputs as follows:

\begin{align}
  \overline{a}(X_r) =\frac{ \sum_{l=1}^{L} {a_l \cdot p_{a_l}(X_r)}}{\sum_{l=1}^{L} {p_{a_l}(X_r)}}.
\end{align}

Secondly, we estimate the born year of each face subtracting the predicted age from the image file creation date. In such case it will be possible to organize the very large albums gathered over the years. In addition, we use the predicted born year as an additional feature with special weight during the cluster analysis in order to partially overcome the known similarity of young babies in a family.

Finally, we implement several tricks in the cluster refinement module (Fig.~\ref{fig:2}). At first, we specially mark the different faces appeared on the same photo. As such faces must be stored in different groups, we additionally perform complete linkage clustering of every facial cluster. The distance matrix specially designed so that the distances between the faces at the same photo are set to the maximum value, which is much larger than the threshold applied when forming flat clusters. Moreover, we assume that the most important clusters should not contain photos/videos made in one day. Hence, we set a certain threshold for a number of days between the earliest and the eldest photo in a cluster in order to disambiguate large quantity of non-interested faces.

\section{Experimental Results}
\label{sect:exper}

The described approach (Fig.~\ref{fig:2}) is implemented in a special software\footnote{\url{https://github.com/HSE-asavchenko/HSE_FaceRec_tf}} using Python language with the Tensorflow and Keras frameworks and the scikit-learn/scipy/numpy libraries. 

\subsection{Facial Clustering}
\label{sect:clustering}
In this subsection we provide experimental study of the proposed system (Fig.~\ref{fig:2}) in facial clustering task for images gathered in unconstrained environments. The identity features extracted by base MobileNet (Fig.~\ref{fig:1}) are compared to the publicly available CNNs suitable for face recognition, namely, the VGGFace (VGGNet-16)~\cite{parkhi2015deep} and the VGGFace2 (ResNet-50)~\cite{cao2018vggface2}. The VGGFace, VGGFace2 and MobileNet extract $D = 4096$, $D = 2048$ and $D = 1024$ non-negative features in the output of ``fc7", ``pool5\_7x7\_s1" and  ``reshape\_1/Mean" layers from 224x224 RGB images, respectively. 

All hierarchical clustering methods from SciPy library are used with the Euclidean ($L_2$) distance between feature vectors. As the centroid and the Ward's linkage showed very poor performance in all cases, we report only results for single, average, complete, weighted and median linkage methods. In addition, we implemented rank-order clustering~\cite{zhu2011rank}, which was specially developed for organizing faces in photo albums. The parameters of all clustering methods were tuned using 10\% of each dataset. We estimate the following clustering metrics with the scikit-learn library: ARI (Adjusted Rand Index), AMI (Adjusted Mutual Information), homogeneity and completeness. In addition, we estimate the average number of extracted clusters $K$ relative to the number of subjects $C$ and the BCubed F-measure. The latter metric is widely applied in various tasks of grouping faces~\cite{GFW,zhang2016clustering}.

In our experiments we used the following testing data.

\begin{itemize}
  \item Subset of LFW (Labeled Faces in the Wild) dataset~\cite{learned2016labeled}, which was involved into the  face identification protocol~\cite{best2014unconstrained}. $C=596$ subjects who have at least two images in the LFW database and at least one video in the YTF (YouTube Faces) database (subjects in YTF are a subset of those in LFW) are used in all clustering methods. 
  \item Gallagher Collection Person Dataset~\cite{Gallagher}, which contains 931 labeled faces with $C=32$ identities in each of the 589 images. As only eyes positions are available in this dataset, we preliminarily detect faces using MTCNN~\cite{zhang2016joint} and choice the subject with the largest intersection of facial region with given eyes region. If the face is not detected we extract s square region with the size chosen as a 1.5-times distance between eyes.
  \item Grouping Faces in the Wild (GFW)~\cite{GFW} with preliminarily detected facial images from 60 real users’ albums from a Chinese social network portal. The size of an album varies from 120 to 3600 faces, with a maximum number of identities of $C=321$.
\end{itemize}

The average values of clustering performance metrics are presented in Table~\ref{table:1}, Table~\ref{table:2} and Table~\ref{table:3} for LFW, Gallagher and GFW datasets, respectively.

\setlength{\tabcolsep}{4pt}
\begin{table}
\begin{center}
\caption{Clustering Results, LFW subset ($C=596$ subjects)}
\label{table:1}
\begin{tabular}{p{0.1\linewidth}llllp{0.14\linewidth}p{0.14\linewidth}p{0.13\linewidth}}
\hline\noalign{\smallskip}
 & & $K/C$ & ARI & AMI & Homogeneity & Completeness & F-measure\\
\noalign{\smallskip}
\hline
\noalign{\smallskip}
& VGGFace & 1.85 & 0.884 & 0.862 & 0.966 & 0.939 & 0.860\\ 
Single & VGGFace2 & 1.22 & 0.993 & 0.969 & 0.995 & 0.986 & 0.967\\ 
& Ours & 2.00 & 0.983 & 0.851 & 0.998 & 0.935 & 0.880\\ \hline
& VGGFace & 1.17 & 0.980 & 0.937 & 0.985 & 0.971 & 0.950\\ 
Average & VGGFace2 & 1.06 & \textbf{0.997} & \textbf{0.987} & 0.998 & \textbf{0.994} & \textbf{0.987}\\ 
& Ours & 1.11 & 0.995 & 0.971 & 0.993 & 0.987 & 0.966\\ \hline
& VGGFace & 0.88 & 0.616 & 0.848 & 0.962 & 0.929 & 0.823\\ 
Complete & VGGFace2 & 0.91 & 0.760 & 0.952 & 0.986 & 0.978 & 0.932\\ 
& Ours & 0.81 & 0.987 & 0.929 & 0.966 & 0.986 & 0.916\\ \hline
& VGGFace & 1.08 & 0.938 & 0.928 & 0.979 & 0.967 & 0.915\\ 
Weighted & VGGFace2 & 1.08 & \textbf{0.997} & 0.982 & 0.998 & 0.992 & 0.983\\ 
& Ours & 1.08 & 0.969 & 0.959 & 0.990 & 0.981 & 0.986\\ \hline
& VGGFace & 2.84 & 0.827 & 0.674 & 0.987 & 0.864 & 0.751\\ 
Median & VGGFace2 & 1.42 & 0.988 & 0.938 & 0.997 & 0.972 & 0.947\\ 
& Ours & 2.73 & 0.932 & 0.724 & \textbf{0.999} & 0.884 & 0.791\\ \hline
Rank-& VGGFace & 0.84 & 0.786 & 0.812 & 0.955 & 0.915 & 0.842\\ 
Order & VGGFace2 & \textbf{0.98} & 0.712 & 0.791 & 0.989 & 0.907 & 0.888\\ 
& Ours & 0.86 & 0.766 & 0.810 & 0.962 & 0.915 & 0.863\\
\hline
\end{tabular}
\end{center}
\end{table}
\setlength{\tabcolsep}{1.4pt}

\setlength{\tabcolsep}{4pt}
\begin{table}
\begin{center}
\caption{Clustering Results, Gallagher dataset ($C=32$ subjects)}
\label{table:2}
\begin{tabular}{p{0.1\linewidth}llllp{0.14\linewidth}p{0.14\linewidth}p{0.13\linewidth}}
\hline\noalign{\smallskip}
 & & $K/C$ & ARI & AMI & Homogeneity & Completeness & F-measure\\
\noalign{\smallskip}
\hline
\noalign{\smallskip}
& VGGFace & 9.13 & 0.601 & 0.435 & 0.966 & 0.555 & 0.662\\ 
Single & VGGFace2 & 2.75 & 0.270 & 0.488 & 0.554 & 0.778 & 0.637\\ 
& Ours & 12.84 & 0.398 & 0.298 & \textbf{1.000} & 0.463 & 0.482\\ \hline
& VGGFace & 1.84 & 0.858 & 0.792 & 0.916 & 0.817 & 0.874\\ 
Average & VGGFace2 & 2.94 & 0.845 & 0.742 & 0.969 & 0.778 & 0.869\\ 
& Ours & 2.03 & \textbf{0.890} & \textbf{0.809} & 0.962 & 0.832 & \textbf{0.897}\\ \hline
& VGGFace & 1.31 & 0.571 & 0.624 & 0.886 & 0.663 & 0.706\\ 
Complete & VGGFace2 & 0.94 & 0.816 & 0.855 & 0.890 & \textbf{0.869} & 0.868\\ 
& Ours & 1.47 & 0.644 & 0.649 & 0.921 & 0.687 & 0.719\\ \hline
& VGGFace & \textbf{0.97} & 0.782 & 0.775 & 0.795 & 0.839 & 0.838\\ 
Weighted & VGGFace2 & 1.63 & 0.607 & 0.730 & 0.876 & 0.760 & 0.763\\ 
& Ours & 1.88 & 0.676 & 0.701 & 0.952 & 0.735 & 0.774\\ \hline
& VGGFace & 9.16 & 0.613 & 0.433 & 0.942 & 0.555 & 0.663\\ 
Median & VGGFace2 & 4.41 & 0.844 & 0.715 & 0.948 & 0.761 & 0.860\\ 
& Ours & 12.38 & 0.439 & 0.324 & 0.960 & 0.482 & 0.531\\ \hline
Rank-& VGGFace & 1.59 & 0.616 & 0.488 & 0.902 & 0.582 & 0.702\\ 
Order & VGGFace2 & 1.94 & 0.605 & 0.463 & 0.961 & 0.566 & 0.682\\ 
& Ours & 3.06 & 0.249 & 0.251 & 0.986 & 0.424 & 0.398\\
\hline
\end{tabular}
\end{center}
\end{table}
\setlength{\tabcolsep}{1.4pt}

\setlength{\tabcolsep}{4pt}
\begin{table}
\begin{center}
\caption{Clustering Results, GFW dataset (in average, $C=46$ subjects) }
\label{table:3}
\begin{tabular}{p{0.1\linewidth}llllp{0.14\linewidth}p{0.14\linewidth}p{0.13\linewidth}}
\hline\noalign{\smallskip}
 & & $K/C$ & ARI & AMI & Homogeneity & Completeness & F-measure\\
\noalign{\smallskip}
\hline
\noalign{\smallskip}
& VGGFace & 4.10 & 0.440 & 0.419 & 0.912 & 0.647 & 0.616\\ 
Single & VGGFace2 & 3.21 & 0.580 & 0.544 & 0.942 & 0.709 & 0.707\\ 
& Ours & 4.19 & 0.492 & 0.441 & 0.961 & 0.655 & 0.636\\ \hline
& VGGFace & 1.42 & 0.565 & 0.632 & 0.860 & 0.751 & 0.713\\ 
Average & VGGFace2 & 1.59 & 0.603 & \textbf{0.663} & 0.934 & 0.761 & 0.746\\ 
& Ours & 1.59 & \textbf{0.609} & 0.658 & 0.917 & \textbf{0.762} & \textbf{0.751}\\ \hline
& VGGFace & \textbf{0.95} & 0.376 & 0.553 & 0.811 & 0.690 & 0.595\\ 
Complete & VGGFace2 & 1.44 & 0.392 & 0.570 & 0.916 & 0.696 & 0.641\\ 
& Ours & 1.28 & 0.381 & 0.564 & 0.886 & 0.693 & 0.626\\ \hline
& VGGFace & 1.20 & 0.464 & 0.597 & 0.839 & 0.726 & 0.662\\ 
Weighted & VGGFace2 & 1.05 & 0.536 & 0.656 & 0.867 & \textbf{0.762} & 0.710\\ 
& Ours & 1.57 & 0.487 & 0.612 & 0.915 & 0.727 & 0.697\\ \hline
& VGGFace & 5.30 & 0.309 & 0.307 & 0.929 & 0.587 & 0.516\\ 
Median & VGGFace2 & 4.20 & 0.412 & 0.422 & 0.929 & 0.639 & 0.742\\ 
& Ours & 6.86 & 0.220 & 0.222 & \textbf{0.994} & 0.552 & 0.411\\ \hline
Rank-& VGGFace & 0.82 & 0.319 & 0.430 & 0.650 & 0.694 & 0.630\\ 
Order & VGGFace2 & 1.53 & 0.367 & 0.471 & 0.937 & 0.649 & 0.641\\ 
& Ours & 1.26 & 0.379 & 0.483 & 0.914 & 0.658 & 0.652\\
\hline
\end{tabular}
\end{center}
\end{table}
\setlength{\tabcolsep}{1.4pt}

The average linkage is the best method according to most of the metrics of cluster analysis. The usage of the rank-order distance~\cite{zhu2011rank} is not appropriate due to rather low performance. Moreover, this distance requires an additional threshold parameter for the cluster-level rank-order distance. Finally, the computational complexity of such clustering is 3-4-times lower when compared to other hierarchical agglomerative clustering methods. One of the most important conclusion here is that the trained MobileNet (Fig.~\ref{fig:1}) is in most cases more accurate than the widely-used VGGFace. As expected, the quality of our model is slightly lower when compared to the deep ResNet-50 CNN trained on the same VGGFace2 dataset. Surprisingly, the highest BCubed F-measure for the most complex GFW dataset (0.751) is achieved by our model. This value is slightly higher than the best BCubed F-measure (0.745) reported in the original paper~\cite{GFW}. However, the most important advantages of our model from the practical point of view are excellent run-time/space complexity. For example, the inference in our model is 5-10-times faster when compared to the VGGFace and VGGFace2. Moreover, the dimensionality of the feature vector is 2-4-times lower leading to the faster computation of a distance matrix in a clustering method. In addition, our model makes it possible to simultaneously predict age and gender of observed facial image. The next subsection will support this claim. 

\subsection{Video-Based Age and Gender Recognition}
\label{sect:AgeGen}

In this subsection the proposed model is compared with publicly available ConvNets for age/gender prediction:
\begin{itemize}
\item Deep expectation (DEX) VGG16 network trained on the IMDB-Wiki dataset~\cite{rothe2015dex} 
\item Wide ResNet~\cite{zagoruyko2016wide} (weights.28-3.73)\footnote{\url{https://github.com/yu4u/age-gender-estimation}}
\item Wide ResNet (weights.18-4.06) \footnote{\url{https://github.com/Tony607/Keras_age_gender}}
\item FaceNet~\cite{schroff2015facenet} \footnote{\url{https://github.com/BoyuanJiang/Age-Gender-Estimate-TF}}
\item BKNetStyle2 \footnote{\url{https://github.com/truongnmt/multi-task-learning}}
\item SSRNet~\cite{yang2018ssr}\footnote{\url{https://github.com/shamangary/SSR-Net}}
\item MobileNet v2 (Agegendernet) \footnote{\label{Agegendernet}\url{https://github.com/dandynaufaldi/Agendernet}}
\item Two models from InsightFace~\cite{deng2018arcface}: original ResNet-50 and ``new" fast ConvNet\footnote{\url{https://github.com/deepinsight/InsightFace/}}
\item Inception v3 \footnote{\url{https://github.com/dpressel/rude-carnie}} fine-tuned on the Adience dataset~\cite{eidinger2014age}
\item Age\_net/gender\_net~\cite{levi2015age} trained on the Adience dataset~\cite{eidinger2014age}.
\end{itemize}

In contrast to the proposed approach, all these models have been trained only on specific datasets with age and gender labels, i.e., they are fine-tuned from traditional ConvNets pre-trained on ImageNet-1000 and do not use large-scale face recognition datasets.

In addition, several special cases of the MobileNet-based model (Fig.~\ref{fig:1}) were examined. Firstly, we compressed this model using standard Tensorflow quantization graph transforms. Secondly, \textit{all} layers of the proposed model were fine-tuned for age and gender predictions (hereinafter ``Proposed MobileNet, fine-tuned"). Though such tuning obviously reduce the accuracy of face identification with identity features at the output of the base MobileNet, it caused an above-mentioned increase of validation accuracies for gender and age classification Thirdly, in order to compare the proposed multi-output neural network (Fig.~\ref{fig:1}) with conventional approach, we additionally used the same MobileNet-based network but with a single head, which was pre-trained on the same VGGFace2 face recognition problem and then fine-tuned for one task (age or gender recognition), i.e., there are two different models (hereinafter ``MobileNet with single head") for all these tasks. Finally, it was decided to measure the influence of pre-training on face recognition task. Hence, the model (Fig.~\ref{fig:1}) was fine-tuned using the above-mentioned procedure and the same training set with labeled age and gender, but the base MobileNet was pre-trained on conventional ImageNet-1000 dataset rather than on VGGFace2~\cite{cao2018vggface2}. Though such network (hereinafter ``Proposed MobileNet, fine-tuned from ImageNet") cannot be used in facial clustering, it can be applied in gender and age prediction tasks.

I run the experiments on the MSI GP63 8RE laptop (CPU: 4xIntel Core i7 2.2 GHz, RAM: 16 GB, GPU: nVidia GeForce GTX 1060) and two mobile phones, namely: 1) Honor 6C Pro (CPU: MT6750 4x1 GHz and 4x2.5 GHz, RAM: 3 GB); and 3) Samsung S9+ (CPU: 4x2.7 GHz Mongoose M3 and 4x1.8 GHz Cortex-A55, RAM: 6 GB). The size of the model file and average inference time for one facial image are presented in Table~\ref{table:6}. 

\begin{table}[ht]
\centering
\begin{tabular}{lllll}
\noalign{\smallskip}
& Model size, & \multicolumn{3}{c}{Average CPU inference time, ms.}\\
ConvNet & MB & Laptop & Mobile phone 1 & Mobile phone 2\\
\noalign{\smallskip}
\hline
\noalign{\smallskip}
age\_net/gender\_net & 43.75 & 91 & 1082 & 224 \\
DEX & 513.82 & 210 & 2730 & 745 \\
Proposed MobileNet & 13.78 & 21 & 354 & 69 \\
Proposed MobileNet, quantized & 3.41 & 19 & 388 & 61 \\
\end{tabular}
\caption{Performance analysis of ConvNets}
\label{table:6}
\end{table}

As expected, the MobileNets are several times faster than the deeper convolutional networks and require less memory to store their weights. Though the quantization reduces the model size in 4 times, it does not decrease the inference time. Finally, though the computing time for the laptop is significantly lower when compared to the inference on mobile phones, their modern models (``Mobile phone 2") became all the more suitable for offline image recognition. In fact, the proposed model requires only 60-70 ms to extract facial identity features and predict both age and gender, which makes it possible to run complex analytics of facial albums on device.

In the next experiments the accuracy of the proposed models were estimated in gender recognition and age prediction tasks. At first, we deal with University of Tennessee, Knoxville Face Dataset (UTKFace)~\cite{zhang2017age}. The images from complete (``In the Wild") set were pre-processed using the following procedure from the above-mentioned Agegendernet resource\footnoteref{Agegendernet}: faces are detected and aligned with margin 0.4 using get\_face\_chip function from DLib. Next, all images which has no single face detected, are removed. The rest 23060 images are scaled to the same size 224x224. In order to estimate the accuracy of age prediction, eight age ranges from the Adience dataset~\cite{eidinger2014age} were used. If the predicted and real age are included into the same range, then the recognition is assumed to be correct. The results are shown in Table~\ref{table:7}. In contrast to the previous experiment (Table~\ref{table:6}), here the inference time is measured on the laptop's GPU. 

In this experiment the proposed ConvNets (three last lines in Table~\ref{table:7}) have higher accuracy of age/gender recognition when compared to the available models trained only on specific datasets, e.g. IMDB-Wiki or Adience. For example, the best fine-tuned MobileNet provided 2.5-40\% higher accuracy of gender classification. The gain in age prediction performance is even more noticeable: we obtain 1.5-10 years less MAE (mean absolute error) and 10-40\% higher accuracy. Though the gender recognition accuracy of a ConvNet with single head is rather high, multi-task learning causes a noticeable decrease in age prediction quality (up to 0.5 and 4.5\% differences in MAE accuracy). Hence, the proposed approach is definitely more preferable if both age and gender recognition tasks are solved due to the twice-lower running time when compared to the usage of separate age and gender networks. It is interesting that though there exist models with lower size, e.g., SSRNet~\cite{yang2018ssr} or new InsightFace-based model~\cite{deng2018arcface}, the proposed ConvNet provides the fastest inference, which can be explained by special optimization of hardware for such widely used architectures as MobileNet.

\begin{table}[h!]
\centering
\begin{tabular}{p{0.41\linewidth}p{0.1\linewidth}p{0.05\linewidth}p{0.1\linewidth}p{0.1\linewidth}p{0.1\linewidth}}
\noalign{\smallskip}
Models & Gender accuracy, \% & Age MAE & Age accuracy, \% & Model size, Mb & Inference time, ms \\
\noalign{\smallskip}
\hline
\noalign{\smallskip}
DEX & 91.05 & 6.48 & 51.77& 1050.5 & 47.1 \\
Wide ResNet (weights.28-3.73) & 88.12 & 9.07 & 46.27 & 191.2 & 10.5\\
Wide ResNet (weights.18-4.06) & 85.35 & 10.05 & 43.23 & 191.2 & 10.5\\
FaceNet & 89.54 & 8.58 & 49.02 & 89.1 & 20.3 \\
BKNetStyle2 & 57.76 & 15.94 & 23.49 & 29.1 & 12.5 \\
SSRNet & 85.69 & 11.90 & 34.86 & 0.6 & 6.6 \\
MobileNet v2 (Agegendernet) & 91.47 & 7.29 & 53.30 & 28.4 & 11.4\\
ResNet-50 from InsightFace & 87.52 & 8.57 & 48.92 & 240.7 & 25.3\\
``New" model from InsightFace & 84.69 & 8.44 & 48.41 & 1.1 & 5.1 \\
Inception trained on Adience & 71.77 & - & 32.09& 85.4 & 37.7 \\
age\_net/gender\_net & 87.32 & - & 45.07& 87.5 & 8.6 \\
MobileNets with single head & 93.59 & 5.94 & 60.29 & 25.7 & 7.2\\
Proposed MobileNet, fine-tuned from ImageNet & 91.81 & 5.88 & 58.47& 13.8 & 4.7\\
Proposed MobileNet, pre-trained on VGGFace2 & 93.79 & 5.74 & 62.67& 13.8 & 4.7 \\
Proposed MobileNet, fine-tuned & 94.10 & 5.44 & 63.97& 13.8 & 4.7\\
\end{tabular}
\caption{Age and gender recognition results for preprocessed UTKFace (In the Wild) dataset}
\label{table:7}
\end{table}

\begin{table}[h!]
\centering
\begin{tabular}{llll}
\noalign{\smallskip}
Models & Gender accuracy, \% & Age MAE & Age accuracy, \% \\
\noalign{\smallskip}
\hline
\noalign{\smallskip}
DEX & 83.16 & 9.84 & 41.22 \\
Wide ResNet (weights.28-3.73) & 73.01 & 14.07 & 29.32 \\
Wide ResNet (weights.18-4.06) & 69.34 & 13.57 & 37.23\\
FaceNet & 86.14 & 9.60 & 44.70 \\
BKNetStyle2 & 60.93 & 15.36 & 21.63 \\
SSRNet & 72.29 & 14.18 & 30.56 \\
MobileNet v2 (Agegendernet) & 86.12 & 11.21 & 42.02 \\
ResNet-50 from InsightFace & 81.15 & 9.53 & 45.30 \\
``New" model from InsightFace & 80.55 & 8.51 & 48.53\\
Inception trained on Adience & 65.89 & - & 27.01 \\
age\_net/gender\_net & 82.36 & - & 34.18 \\
MobileNets with single head & 91.89 & 6.73 & 57.21\\
Proposed MobileNet, fine-tuned from ImageNet & 84.30 & 7.24 & 58.05 \\
Proposed MobileNet, pre-trained on VGGFace2 & 91.95 & 6.00 & 61.70 \\
Proposed MobileNet, fine-tuned & 91.95 & 5.96 & 62.74 \\
\end{tabular}
\caption{Age and gender recognition results for UTKFace (aligned \& cropped faces) dataset}
\label{table:8}
\end{table}

It is known that various image preprocessing algorithms could drastically influence the recognition performance. Hence, in the next experiment conventional set of all 23708 images from ``Aligned \& cropped faces" provided by the authors of the UTKFace was used. The results of the same models are presented in Table~\ref{table:8}.

The difference in image pre-processing causes significant \textit{decrease} of the accuracies of most models. For example, the best proposed model in this experiment has 14-40\% and 5-40\% higher age and gender recognition accuracy, respectively. Its age prediction MAE is at least 2.5 years lower when compared to the best publicly available model from Insight face. The usage of the same DLib library to detect and align faces in a way, which is only slightly different from the above-mentioned preprocessing pipeline, drastically decreases the accuracy of the best existing model from previous experiment (MobileNet v2) up to 5.5\% gender accuracy and 3 years in age prediction MAE. Obviously, such dependance of performance on the image pre-processing algorithm makes the model inappropriate for practical applications. Hence, this experiment clearly demonstrates how the proposed model exploits the potential of very large face recognition dataset to learn face representation in contrast to the training only on rather small and dirty datasets with age and gender labels. It is important to emphasize that the same statement is valid even for the proposed model (Fig.~\ref{fig:1}). In particular, the usage of face identification features pre-trained on VGGFace2 leads to 3.5\% and 6.5\% lower error rate of age and gender classification, respectively, when compared to conventional fine-tuning of MobileNet, which has been preliminarily trained on ImageNet-1000 only (third last line in Table~\ref{table:8}). This difference in error rates is much more noticeable when compared to the previous experiment (Table~\ref{table:7}), especially for age prediction MAE.

Many recent papers devoted to UTKFace dataset split it into the training and testing sets and fine-tune the models on such training set~\cite{das2018mitigating}. Though the proposed ConvNet does not require such fine-tuning, its results are still very competitive. For example, we used the testing set described in the paper~\cite{cao2019consistent}, which achieves the-state-of-the-art results on a subset of UTKFace if only 3287 photos of persons from the age ranges [21, 60] are taken into the testing set. The proposed model achieves 97.5\% gender recognition accuracy and age prediction MAE 5.39. It is lower than 5.47 MAE of the best CORAL-CNN model from this paper, which was additionally trained on other subset of UTKFace. 

As the age and gender recognition is performed in the proposed pipeline (Fig.~\ref{fig:2}) for a \textit{set} of facial images in a cluster, it was decided to use the known video datasets with age/gender labels in the next experiments in order to test performance of classification of a set of video frames:

\begin{itemize}
 \item Eurecom Kinect~\cite{min2014kinectfacedb}, which contains 9 photos for each of 52 subjects (14 women and 38 men). 
 \item Indian Movie Face database (IMFDB)~\cite{setty2013indian} with 332 video clips of 63 males and 33 females. Only four age categories are available: ``Child" (0-15 years old), ``Young" (16-35), ``Middle: (36-60) and ``Old" (60+).
 \item Acted Facial Expressions in the Wild (AFEW) from the EmotiW 2018 (Emotions recognition in the wild) audio-video emotional sub-challenge~\cite{klare2015pushing}. It contains 1165 video files. The facial regions were detected using the MTCNN~\cite{zhang2016joint}.
 \item IARPA Janus Benchmark A (IJB-A)~\cite{dhall2012collecting} with more than 13000 total frames of 1165 video tracks. Only gender information is available in this dataset.
 \end{itemize}

In video-based gender recognition a gender of a person on each video frame is firstly classified. After that two simple fusion strategies are utilized, namely, simple voting and the product rule (1). The obtained accuracies of the proposed models compared to most widely used DEX~\cite{rothe2015dex} and gender\_net/age\_net~\cite{levi2015age} are shown in Table~\ref{table:9}.

\begin{table}[h!]
\centering
\begin{tabular}{llllll}
\noalign{\smallskip}
 ConvNet & Aggregation & Eurecom Kinect & IMFDB & AFEW & IJB-A\\
\noalign{\smallskip}
\hline
\noalign{\smallskip}
gender\_net & Simple Voting & 73 & 71 & 75 & 60 \\
& Product rule & 77 & 75 & 75 & 59 \\
DEX & Simple Voting & 84 & 81 & 80 & 81 \\
& Product rule & 84 & 88 & 81 & 82 \\
MobileNet with & Simple Voting & 93 & 97 & 92 & 95 \\
single head& Product rule & 93 & 98 & \textbf{93} & 95\\
Proposed MobileNet & Simple Voting & 94 & 98 & \textbf{93} & 95 \\
& Product rule & 93 & \textbf{99} & \textbf{93} & \textbf{96}\\
Proposed MobileNet, & Simple Voting & 88 & 96 & 92 & 93\\
quantized & Product rule & 86 & 96 & 93 & 94 \\
Proposed MobileNet, & Simple Voting & 93 & 95 & 91 & 94\\
fine-tuned & Product rule & \textbf{95} & 97 & 92 & 95\\
\end{tabular}
\caption{Video-based gender recognition accuracy, \%}
\label{table:9}
\end{table}

\begin{table}[h!]
\centering
\begin{tabular}{lllll}
\noalign{\smallskip}
 ConvNet & Aggregation & Eurecom Kinect & IMFDB & AFEW \\
\noalign{\smallskip}
\hline
\noalign{\smallskip}
 & Simple Voting & 41 & 68 & 27 \\
age\_net & Product Rule & 45 & 48 & 27 \\
& Expected Value & 69 & 32 & 30 \\ \hline
 & Simple Voting & 60 & 29 & 47 \\
DEX & Product Rule & 71 & 29 & 48 \\
& Expected Value & 71 & 54 & 52 \\ \hline
MobileNet with & Simple Voting & 91 & 34 & 46 \\
single head& Product Rule & 93 & 38 & 47 \\
& Expected Value & \textbf{94} & 75 & \textbf{54}\\ \hline
Proposed MobileNet, & Simple Voting & 73 & 30 & 47 \\
IMDB-Wiki only& Product Rule & 83 & 31 & 47 \\
& Expected Value & 85 & 58 & 52\\ \hline
Proposed MobileNet, & Simple Voting & 92 & 32 & 46 \\
IMDB-Wiki+Adience& Product Rule & \textbf{94} & 36 & 46 \\
& Expected Value & \textbf{94} & \textbf{77} & \textbf{54}\\ \hline
Proposed MobileNet, & Simple Voting & 86 & 34 & 44\\
quantized & Product Rule & 88 & 36 & 46\\
 & Expected Value & 85 & 58 & 50 \\ \hline
Proposed MobileNet, & Simple Voting & 74 & 33 & 45\\
fine-tuned & Product Rule & 77 & 35 & 45\\
 & Expected Value & 92 & 72 & 51\\
\end{tabular}
\caption{Video-based age prediction accuracy, \%}
\label{table:10}
\end{table}

Here again the proposed models are much more accurate than the publicly available ConvNets trained only on rather small and dirty datasets with age/gender labels. It is important to emphasize that the gender output of the proposed network was trained on the same IMDB-Wiki dataset as the DEX network~\cite{rothe2015dex}. However, the error rate in the proposed approach is much lower when compared to the DEX. It can be explained by the pre-training of the base MobileNet on face identification task with very large dataset, which helps to learn rather good facial representations. Such pre-training differs from traditional usage of ImageNet weights and only fine-tune the CNN on a specific dataset with known age and gender labels. Secondly, the usage of product rule generally leads to 1-2\% decrease of the error rate when compared to the simple voting. Thirdly, the fine-tuned version of the proposed model achieves the lowest error rate only for the Kinect dataset and is 1-3\% \textit{less} accurate in other cases. It is worth noting that the best accuracy for Eurecom Kinect dataset is 7\% higher than the best known accuracy (87.82\%) achieved by~\cite{huynh2012efficient} in similar settings without analysis of depth images. Fourthly, though the compression of the ConvNet makes it possible to drastically reduce the model size (Table~\ref{table:6}), it is characterized by up to 7\% decrease of the recognition rate. Finally, conventional single-output model is slightly less accurate than the proposed network, though the difference in the error rate is not statistically significant. 

In the last experiment the results for age predictions are presented (Table~\ref{table:10}). As the training set for the proposed network differs with conventional DEX model due to addition of the Adience data to the IMDB-Wiki dataset, it was decided to repeat training of the proposed network (Fig.~\ref{fig:1}) with the IMDB-Wiki data only. Hence, the resulted ``Proposed MobileNet, IMDB-Wiki only" ConvNet is more fairly compared with the DEX network.

Here it was assumed that age is recognized correctly for the Kinect and AFEW datasets (with known age) if the difference between real and predicted age is not greater than 5 years. The fusion of age predictions of individual video frames is implemented by: 1) simple voting, 2) maximizing the product of age posterior probabilities (2), and 3) averaging the expected value (3) with choice of $L=3$ top predictions in each frame. 

One can notice that the proposed models are again the most accurate in practically all cases. The accuracy of the DEX models are comparable with the proposed ConvNets only for the AFEW dataset. This gain in the error rate cannot be explained by using additional Adience data, as it is noticed even for the ``Proposed MobileNet, IMDB-Wiki only" model. Secondly, the lowest error rates are obtained for the computation of the expected value of age predictions. For example, it is 2\% and 8\% more accurate than the simple voting for the Kinect and AFEW data. The effect is especially clear for the IMFDB images, in which the expected value leads to up to 45\% higher recognition rate. 

\section{Conclusions}
\label{sect:conclusion}

In this paper we proposed an approach to organizing photo and video albums (Fig.~\ref{fig:2}) based on a simple extension of MobileNet (Fig.~\ref{fig:1}), in which the facial representations suitable for face identification, age and gender recognition problems are extracted. The main advantage of the proposed model is the possibility to solve all three tasks simultaneously without need for additional ConvNets. As a result, a very fast facial analytic system  was implemented (Table~\ref{table:6}), which can be installed even on mobile devices. It was shown that the proposed approach extracts facial clusters rather accurately when compared to the known models (Table~\ref{table:1} and Table~\ref{table:2}). Moreover, the known state-of-the-art BCubed F-measure for very complex GFW data was slightly improved (Table~\ref{table:3}). What is more important, the results for age and gender predictions using extracted facial representations significantly outperform the results of the publicly available neural networks (Table~\ref{table:9} and Table~\ref{table:10}). The state-of-the-art results on the whole UTKFace dataset~\cite{zhang2017age} was achieved (94.1\% gender recognition accuracy, 5.44 age prediction MAE) for the ConvNets which are not fine-tuned on a part of this dataset.

The proposed approach does not have limitations of existing methods of simultaneous face analysis~\cite{ranjan2017hyperface,yoo2018method} for usage in face identification and clustering tasks because it firstly learns the facial representations using external very large face recognition dataset. The proposed approach is usable even for face identification with small training samples, including the most complex case, namely, a single photo per individual. Furthermore, the method enables to apply the ConvNet in completely unsupervised environments for face clustering, given only a set of photos from a mobile device. Finally, the training procedure to learn parameters of the method alternately trains all the facial attribute recognition tasks using batches of different training images. Hence, the training images are not required to have all attributes available. As a result, much more complex (but still fast) network architectures can be trained when compared to the ConvNet of~\cite{yoo2018method} and, hence, achieve very high age/gender recognition accuracy and the face clustering quality comparable to very deep state-of-the-art ConvNets. 

In future works it is necessary to deal with the aging problem. In fact, the average linkage clustering usually produces several groups for the same person (especially, a child). The single linkage clustering can resolve this issue if there exist photos of the same subject made over the years. Unfortunately, the performance of the single linkage is rather poor when compared to another agglomerative clustering methods (Table~\ref{table:1}, Table~\ref{table:2} and Table~\ref{table:3}). An additional research direction is a thorough analysis of distance measures in the facial clustering~\cite{zhu2011rank}, i.e., the usage of distance learning~\cite{zhu2013point} or special regulizers~\cite{savchenko2018unconstrained}. Finally, it is necessary to examine more complex aggregation techniques, i.e., learnable pooling~\cite{miech2017learnable} or special implementation of attention mechanism~\cite{yang2017neural} in order to improve the quality of decision-making based on all facial images in a cluster.

\section*{Acknowledgment}
This research was supported by Samsung Research and Samsung Electronics. Additionally, this research was prepared within the framework of the Basic Research Program at the National Research University Higher School of Economics (HSE) in return for lab time.

\bibliographystyle{splncs}
\bibliography{egbib}

\begin{thebibliography}{10}

\bibitem{manju2015organizing}
Manju, A., Valarmathie, P.:
\newblock Organizing multimedia big data using semantic based video content
  extraction technique.
\newblock In: Soft-Computing and Networks Security (ICSNS), 2015 International
  Conference on, IEEE (2015)  1--4

\bibitem{sokolova2017organizing}
Sokolova, A.D., Kharchevnikova, A.S., Savchenko, A.V.:
\newblock Organizing multimedia data in video surveillance systems based on
  face verification with convolutional neural networks.
\newblock In: International Conference on Analysis of Images, Social Networks
  and Texts, Springer (2017)  223--230

\bibitem{zhang2002hierarchical}
Zhang, Y.J., Lu, H.:
\newblock A hierarchical organization scheme for video data.
\newblock Pattern Recognition \textbf{35}(11) (2002)  2381--2387

\bibitem{GFW}
He, Y., Cao, K., Li, C., Loy, C.C.:
\newblock Merge or not? {Learning} to group faces via imitation learning.
\newblock arXiv preprint arXiv:1707.03986 (2017)

\bibitem{eidinger2014age}
Eidinger, E., Enbar, R., Hassner, T.:
\newblock Age and gender estimation of unfiltered faces.
\newblock IEEE Transactions on Information Forensics and Security
  \textbf{9}(12) (2014)  2170--2179

\bibitem{rothe2015dex}
Rothe, R., Timofte, R., Van~Gool, L.:
\newblock {DEX}: Deep expectation of apparent age from a single image.
\newblock In: Proceedings of the IEEE International Conference on Computer
  Vision Workshops. (2015)  10--15

\bibitem{goodfellow2016deep}
Goodfellow, I., Bengio, Y., Courville, A.:
\newblock Deep learning.
\newblock MIT press (2016)

\bibitem{crosswhite2017template}
Crosswhite, N., Byrne, J., Stauffer, C., Parkhi, O., Cao, Q., Zisserman, A.:
\newblock Template adaptation for face verification and identification.
\newblock In: Automatic Face \& Gesture Recognition (FG 2017), 2017 12th IEEE
  International Conference on, IEEE (2017)  1--8

\bibitem{wang2018additive}
Wang, F., Cheng, J., Liu, W., Liu, H.:
\newblock Additive margin softmax for face verification.
\newblock IEEE Signal Processing Letters \textbf{25}(7) (2018)  926--930

\bibitem{savchenko2018unconstrained}
Savchenko, A.V., Belova, N.S.:
\newblock Unconstrained face identification using maximum likelihood of
  distances between deep off-the-shelf features.
\newblock Expert Systems with Applications \textbf{108} (2018)  170--182

\bibitem{ranjan2017hyperface}
Ranjan, R., Patel, V.M., Chellappa, R.:
\newblock Hyperface: A deep multi-task learning framework for face detection,
  landmark localization, pose estimation, and gender recognition.
\newblock IEEE Transactions on Pattern Analysis and Machine Intelligence (2017)

\bibitem{howard2017mobilenets}
Howard, A.G., Zhu, M., Chen, B., Kalenichenko, D., Wang, W., Weyand, T.,
  Andreetto, M., Adam, H.:
\newblock {MobileNets}: Efficient convolutional neural networks for mobile
  vision applications.
\newblock arXiv preprint arXiv:1704.04861 (2017)

\bibitem{cao2018vggface2}
Cao, Q., Shen, L., Xie, W., Parkhi, O.M., Zisserman, A.:
\newblock {VGGFace2}: A dataset for recognising faces across pose and age.
\newblock In: Automatic Face \& Gesture Recognition (FG 2018), 2018 13th IEEE
  International Conference on, IEEE (2018)  67--74

\bibitem{zhang2016joint}
Zhang, K., Zhang, Z., Li, Z., Qiao, Y.:
\newblock Joint face detection and alignment using multitask cascaded
  convolutional networks.
\newblock IEEE Signal Processing Letters \textbf{23}(10) (2016)  1499--1503

\bibitem{aggarwal2013data}
Aggarwal, C.C., Reddy, C.K.:
\newblock Data clustering: algorithms and applications.
\newblock CRC press (2013)

\bibitem{parkhi2015deep}
Parkhi, O.M., Vedaldi, A., Zisserman, A.,  et~al.:
\newblock Deep face recognition.
\newblock In: BMVC. Volume~1. (2015) ~6

\bibitem{kaya2017video}
Kaya, H., G{\"u}rp{\i}nar, F., Salah, A.A.:
\newblock Video-based emotion recognition in the wild using deep transfer
  learning and score fusion.
\newblock Image and Vision Computing \textbf{65} (2017)  66--75

\bibitem{rassadin2017group}
Rassadin, A., Gruzdev, A., Savchenko, A.:
\newblock Group-level emotion recognition using transfer learning from face
  identification.
\newblock In: Proceedings of the 19th ACM International Conference on
  Multimodal Interaction, ACM (2017)  544--548

\bibitem{kittler1998combining}
Kittler, J., Hatef, M., Duin, R.P., Matas, J.:
\newblock On combining classifiers.
\newblock IEEE Transactions on Pattern Analysis and Machine Intelligence
  \textbf{20}(3) (1998)  226--239

\bibitem{zhu2011rank}
Zhu, C., Wen, F., Sun, J.:
\newblock A rank-order distance based clustering algorithm for face tagging.
\newblock In: Computer Vision and Pattern Recognition (CVPR), 2011 IEEE
  Conference on, IEEE (2011)  481--488

\bibitem{zhang2016clustering}
Zhang, Z., Luo, P., Loy, C.C., Tang, X.:
\newblock Joint face representation adaptation and clustering in videos.
\newblock In: European conference on computer vision, Springer (2016)  236--251

\bibitem{learned2016labeled}
Learned-Miller, E., Huang, G.B., RoyChowdhury, A., Li, H., Hua, G.:
\newblock Labeled faces in the wild: A survey.
\newblock In: Advances in face detection and facial image analysis.
\newblock Springer (2016)  189--248

\bibitem{best2014unconstrained}
Best-Rowden, L., Han, H., Otto, C., Klare, B.F., Jain, A.K.:
\newblock Unconstrained face recognition: Identifying a person of interest from
  a media collection.
\newblock IEEE Transactions on Information Forensics and Security
  \textbf{9}(12) (2014)  2144--2157

\bibitem{Gallagher}
Gallagher, A.C., Chen, T.:
\newblock Clothing cosegmentation for recognizing people.
\newblock In: Computer Vision and Pattern Recognition, 2008. CVPR 2008. IEEE
  Conference on, IEEE (2008)  1--8

\bibitem{zagoruyko2016wide}
Zagoruyko, S., Komodakis, N.:
\newblock Wide residual networks.
\newblock arXiv preprint arXiv:1605.07146 (2016)

\bibitem{schroff2015facenet}
Schroff, F., Kalenichenko, D., Philbin, J.:
\newblock {FaceNet}: A unified embedding for face recognition and clustering.
\newblock In: Proceedings of the IEEE Conference on Computer Vision and Pattern
  Recognition (CVPR. (2015)  815--823

\bibitem{yang2018ssr}
Yang, T.Y., Huang, Y.H., Lin, Y.Y., Hsiu, P.C., Chuang, Y.Y.:
\newblock Ssr-net: A compact soft stagewise regression network for age
  estimation.
\newblock In: Proceedings of the International Joint Conference on Artificial
  Intelligence (IJCAI). (2018)  1078--1084

\bibitem{deng2018arcface}
Deng, J., Guo, J., Niannan, X., Zafeiriou, S.:
\newblock Arcface: Additive angular margin loss for deep face recognition.
\newblock In: Proceedings of the IEEE Conference on Computer Vision and Pattern
  Recognition (CVPR). (2019)

\bibitem{levi2015age}
Levi, G., Hassner, T.:
\newblock Age and gender classification using convolutional neural networks.
\newblock In: Proceedings of the IEEE Conference on Computer Vision and Pattern
  Recognition Workshops. (2015)  34--42

\bibitem{zhang2017age}
Zhang, Z., Song, Y., Qi, H.:
\newblock Age progression/regression by conditional adversarial autoencoder.
\newblock In: Proceedings of the IEEE Conference on Computer Vision and Pattern
  Recognition (CVPR). (2017)  5810--5818

\bibitem{das2018mitigating}
Das, A., Dantcheva, A., Bremond, F.:
\newblock Mitigating bias in gender, age and ethnicity classification: A
  multi-task convolution neural network approach.
\newblock In: European Conference on Computer Vision, Springer (2018)  573--585

\bibitem{cao2019consistent}
Cao, W., Mirjalili, V., Raschka, S.:
\newblock Consistent rank logits for ordinal regression with convolutional
  neural networks.
\newblock arXiv preprint arXiv:1901.07884 (2019)

\bibitem{min2014kinectfacedb}
Min, R., Kose, N., Dugelay, J.L.:
\newblock Kinectfacedb: A {Kinect} database for face recognition.
\newblock IEEE Transactions on Systems, Man, and Cybernetics: Systems
  \textbf{44}(11) (2014)  1534--1548

\bibitem{setty2013indian}
Setty, S., Husain, M., Beham, P., Gudavalli, J., Kandasamy, M., Vaddi, R.,
  Hemadri, V., Karure, J., Raju, R., Rajan, B.,  et~al.:
\newblock Indian movie face database: a benchmark for face recognition under
  wide variations.
\newblock In: Fourth National Conference on Computer Vision, Pattern
  Recognition, Image Processing and Graphics (NCVPRIPG), IEEE (2013)  1--5

\bibitem{klare2015pushing}
Klare, B.F., Klein, B., Taborsky, E., Blanton, A., Cheney, J., Allen, K.,
  Grother, P., Mah, A., Jain, A.K.:
\newblock Pushing the frontiers of unconstrained face detection and
  recognition: Iarpa janus benchmark a.
\newblock In: Proceedings of the IEEE Conference on Computer Vision and Pattern
  Recognition. (2015)  1931--1939

\bibitem{dhall2012collecting}
Dhall, A.,  et~al.:
\newblock Collecting large, richly annotated facial-expression databases from
  movies.
\newblock IEEE Multimedia (2012)

\bibitem{huynh2012efficient}
Huynh, T., Min, R., Dugelay, J.L.:
\newblock An efficient {LBP}-based descriptor for facial depth images applied
  to gender recognition using {RGB-D} face data.
\newblock In: Asian Conference on Computer Vision, Springer (2012)  133--145

\bibitem{yoo2018method}
Yoo, B., Namjoon, K., Changkyo, L., Choi, C.K., JaeJoon, H.:
\newblock Method and apparatus for recognizing object, and method and apparatus
  for training recognizer (March~27 2018) US Patent 9,928,410.

\bibitem{zhu2013point}
Zhu, P., Zhang, L., Zuo, W., Zhang, D.:
\newblock From point to set: Extend the learning of distance metrics.
\newblock In: Proceedings of the International Conference on Computer Vision
  (ICCV), IEEE (2013)  2664--2671

\bibitem{miech2017learnable}
Miech, A., Laptev, I., Sivic, J.:
\newblock Learnable pooling with context gating for video classification.
\newblock arXiv preprint arXiv:1706.06905 (2017)

\bibitem{yang2017neural}
Yang, J., Ren, P., Chen, D., Wen, F., Li, H., Hua, G.:
\newblock Neural aggregation network for video face recognition.
\newblock In: Proceedings of the International Conference on Computer Vision
  and Pattern Recognition (CVPR), IEEE (2017)  4362--4371

\end{thebibliography}
\end{document}